\documentclass[a4paper,11pt,twocolumn,twoside]{article}
\usepackage{graphicx}
\usepackage{sepln_en}
\usepackage{fullname}
\usepackage[utf8]{inputenc}
\usepackage[spanish, english]{babel}
\usepackage{xcolor}
\usepackage{url}
\usepackage{multirow}
\usepackage{svg}

\title{Bertinho: Galician BERT Representations} 

\author {\textbf{David Vilares,$^1$} \textbf{Marcos Garcia,$^2$} \textbf{Carlos Gómez-Rodríguez\,$^1$}\\
$^1$Universidade da Coruña, CITIC, Galicia, Spain\\
$^2$CiTIUS, Universidade de Santiago de Compostela, Galicia, Spain\\
david.vilares@udc.es, marcos.garcia.gonzalez@usc.gal, carlos.gomez@udc.es
}
\seplntranstitle{Bertinho: Representaciones BERT para el gallego}

\seplnclave{BERT, gallego, \emph{embeddings}, modelado del lenguaje.}

\seplnresumen{Este artículo presenta un modelo BERT monolingüe para el gallego. Nos basamos en la tendencia actual que ha demostrado que es posible crear modelos BERT monolingües robustos incluso para aquellos idiomas para los que hay una relativa escasez de recursos, funcionando éstos mejor que el modelo BERT multilingüe oficial (mBERT). Concretamente, liberamos dos modelos monolingües para el gallego, creados con 6 y 12 capas de \emph{transformers}, respectivamente, y entrenados con una limitada cantidad de recursos ($\sim$45 millones de palabras sobre una única GPU de 24GB.)
Para evaluarlos realizamos un conjunto exhaustivo de experimentos en tareas como análisis morfosintáctico, análisis sintáctico de dependencias o reconocimiento de entidades. Para ello, abordamos estas tareas como etiquetado de secuencias, con el objetivo de ejecutar los modelos BERT sin la necesidad de incluir ninguna capa adicional (únicamente se añade la capa de salida encargada de transformar las representaciones contextualizadas en la etiqueta predicha). Los experimentos muestran que nuestros modelos, especialmente el de 12 capas, mejoran los resultados de mBERT en la mayor parte de las tareas.
}

\seplnkey{BERT, Galician, embeddings, language modeling.}

\seplnabstract{This paper presents a monolingual BERT model for Galician. We follow the recent trend that shows that it is feasible to build robust monolingual BERT models even for relatively low-resource languages, while performing better than the well-known official multilingual BERT (mBERT). More particularly, we release two monolingual Galician BERT models, built using 6 and 12 transformer layers, respectively; trained with limited resources ($\sim$45 million tokens on a single GPU of 24GB). We then provide an exhaustive evaluation on a number of tasks such as POS-tagging, dependency parsing and named entity recognition. For this purpose, all these tasks are cast in a pure sequence labeling setup in order to run BERT without the need to include any additional layers on top of it (we only use an output classification layer to map the contextualized representations into the predicted label). The experiments show that our models, especially the 12-layer one, outperform the results of mBERT in most tasks.
}

\firstpageno{1}

\begin{document}



\label{firstpage} \maketitle

\section{Introduction}




Contextualized word representations \cite{dai2015semi,peters-etal-2018-deep,devlin-etal-2019-bert} have largely improved the performance of many natural language processing (NLP) tasks, such as syntactic parsing \cite{kitaev-klein-2018-constituency}, question answering \cite{salant-berant-2018-contextualized} or natural language inference \cite{jiang-de-marneffe-2019-evaluating}, among many others. Contrary to static word embeddings \cite{mikolov2013distributed,pennington-etal-2014-glove}, where a given term is always represented by the same low-dimensional vector, contextualized approaches encode each word based on its context. Such process is normally learned by a neural network that optimizes a language modeling objective.

One of the most popularized and best performing models to generate contextualized representations is BERT \cite{devlin-etal-2019-bert}, a bidirectional language model based on transformers \cite{vaswani2017attention}. BERT was initially released as a monolingual model for English, with \emph{large} and \emph{base} variants, made of 24 and 12 transformer layers, respectively. In addition, a multilingual BERT version (mBERT) trained on the one hundred most popular languages on Wikipedia was also released. Although mBERT has become a very popular and easy-to-use tool to address multilingual NLP challenges \cite{pires-etal-2019-multilingual,K2020Cross-Lingual}, some authors have reported that its performance is not so robust as that of the corresponding monolingual models \cite{wu-dredze-2020-languages,vulic-etal-2020-probing}. In this line, previous work has showed that training a monolingual BERT is worth in terms of performance, in comparison to mBERT. Among others, this is the case for languages coming from different typologies, languages families and scripts, such as Finnish \cite{virtanen2019multilingual}, Basque \cite{agerri-etal-2020-give}, Spanish \cite{CaneteCFP2020}, Greek \cite{Koutsikakis2020} or Korean \cite{lee2020kr}.

Taking the above into account, this paper contributes with the development of BERT models for Galician, a relatively low-resource language for which, to best of our knowledge, there is no contextualized (monolingual) model available. In this regard, we train two \emph{Bertinho} models and test their performance on several tasks. Specifically, we assess the effect of the number of layers when using limited data (less than 45 million tokens). To do so, we train models with 6 and 12 layers using a single TESLA P40 24GB GPU and compare them against the official multilingual BERT
on a number of downstream tasks, including POS-tagging, dependency parsing, and named entity recognition (NER).
The experiments show that the monolingual models clearly outperform mBERT: even a small 6-layer model outperforms the official multilingual one in most scenarios, and the 12-layer one obtains the overall best results. We have submitted \emph{Bertinho} to the HuggingFace Models Hub\footnote{\url{https://huggingface.co/dvilares/bertinho-gl-base-cased} and \url{https://huggingface.co/dvilares/bertinho-gl-small-cased}}. Contemporaneously to this work, HuggingFace has released a Galician RoBERTa \cite{liu2019roberta} model\footnote{\url{https://huggingface.co/mrm8488/RoBERTinha}} based on the approach presented by \namecite{ortiz-suarez-etal-2020-monolingual}.

Apart from this introduction, this paper is organized as follows: Section~\ref{sec:rw} introduces some related work regarding static and contextualized vector representations for NLP, discussing different approaches to train new monolingual models. Then, we describe the particularities of the Galician BERT-based models in Section~\ref{sec:models}. Finally, the experiments and results are presented and discussed in Section~\ref{sec:experiments}, while the conclusions are drawn in Section~\ref{sec:conclusions}.

\section{Related Work\label{sec:rw}}
The paradigm shift of NLP architectures produced by the rise of neural networks \cite{bengio2003neural,collobert2008unified,collobert2011natural} popularized the use of vector space models, following previous work in distributional semantics \cite{landauer1997solution,mcdonald2001testing}. In this scenario, several highly efficient methods have been proposed to learn low-dimensional vector representation of words (i.e., word embeddings), such as \textit{word2vec} \cite{mikolov2013}, \textit{GloVe} \cite{pennington-etal-2014-glove}, or \textit{fastText} \cite{bojanowski-etal-2017-enriching}. Since then, the use of pretrained embeddings to initialize the training of deep learning NLP models has become a standard procedure, due to the positive impact provided by the distributed representations in most downstream tasks \cite{schnabel-etal-2015-evaluation}.

One of the main drawbacks of these \emph{static} word embeddings for NLP is that they represent all the senses of a given word in the same vector, thus making it difficult to deal with different linguistic phenomena such as polysemy or homonymy. In this regard, \namecite{peters-etal-2018-deep} introduced ELMo,
a model which obtains contextualized vector representations by means of an LSTM architecture, thus providing context-specific vectors for each token. This language model runs in a unidirectional fashion, in order not to trick itself when predicting the next word.

Following this idea, \namecite{devlin-etal-2019-bert} presented BERT, a bidirectional language representation model based on the transformer architecture \cite{vaswani2017attention} and trained on a masked language model objective and on a next sentence prediction one, in order to consider both previous and upcoming context while still being a fair training objective. BERT has obtained state-of-the-art results for many NLP tasks, and it is easy executable through freely available hubs, becoming a strong baseline in many recent work. In addition, the success achieved by BERT has promoted an extensive implementation of BERT-based models with several goals, such as optimizing its training method, e.g. RoBERTa, \cite{liu2019roberta} or reducing the size and complexity of the model, e.g. DistilBERT \cite{sanh2020distilbert} or ALBERT \cite{albert}\footnote{For which there is a monolingual Catalan model: \url{https://github.com/codegram/calbert}}. Moreover, it has attracted the interest of the research community regarding how its deep architecture encodes linguistic knowledge \cite{lin-etal-2019-open,VilStrSogGomAAAI2020,ettinger-2020-bert}.  

Besides the original English and Chinese models of BERT, the authors released a multilingual one (mBERT, with 12 layers) which produces inter-linguistic representations to some extent and gives good performance at zero-shot cross-lingual tasks \cite{pires-etal-2019-multilingual}. However, several studies have pointed that there are significant differences in performance among the languages covered by mBERT \cite{wu-dredze-2020-languages}, and that for some tasks even the best represented ones do not produce competitive results \cite{vulic-etal-2020-probing}.
These findings suggest that, if possible, it is worth training monolingual models, especially for those languages which are poorly represented in the multilingual version of BERT.

In fact, monolingual versions of BERT have been trained for various languages using different methods, improving the results of mBERT. Finnish \cite{virtanen2019multilingual}, Spanish \cite{CaneteCFP2020} and Greek \cite{Koutsikakis2020} models were trained with about 3 billion tokens using the same parameters as the original BERT-Base (12 layers and 768 vector dimensions). For Portuguese, \namecite{souza2019portuguese} trained two models using about 2.7 billion tokens: one \emph{large}, using the original English BERT-Large (with 24 layers and a vector size of 1024) for initialization, and one \emph{base}, initialized from mBERT. A similar approach has been also adopted for Russian \cite{yu2019adaptation}, whose RuBERTa model used mBERT as the starting point, too. A comparison with a monolingual training with random initialization showed that starting from the pre-trained mBERT reduces training time and allows for achieving better performance in various tasks.
Finally, and more similar to our setting, the monolingual Basque model \cite{agerri-etal-2020-give} obtains state-of-the-art performance in different downstream tasks using only 225 million tokens for training a BERT-base model. 

With the above in mind, this paper presents the work carried out to train two BERT models for Galician (one small, with only 6 layers, and one base, with 12), and evaluate them in downstream tasks such as POS-tagging, dependency parsing, and named entity recognition. Our approach can be seen as a low-resource scenario, as we only use one GPU and a small corpus of 42 million tokens for training.

\section{Bertinho models\label{sec:models}}
This section briefly introduces some ideas about Galician and describes the methodology that we have followed to train and evaluate the \emph{Bertinho} models.

\subsection{Galician}
Galician is a romance language spoken by about 2.5 million people in the Spanish Autonomous Region of Galicia and adjacent territories \cite{ige}. It belongs to the Western Ibero-Romance group, being evolved from the medieval Galician-Portuguese \cite{teyssier}. Both philological and linguistic studies have traditionally classified Galician dialects as part of the same language as Portuguese \cite{cintra,freixeiro}, even though Galician has been standardized as an independent language since the 1970s, mainly through the use of a Spanish-based orthography \cite{samartim2012}. In this regard, both Spanish and Portuguese NLP resources and tools have been used and adapted to analyse Galician data \cite{malvar2010vencendo,garcia2018new}.

\subsection{Training data for language modeling with BERT\label{sec:data}}
For the pre-training phase, where BERT will be trained for language modeling in order to learn to generate robust contextualized word representations, we rely on a small corpus, extracted from the Galician version of the Wikipedia.
More particularly, we used the 2020-02-01 dump\footnote{Note that only the newest dumps are maintained over time \url{https://dumps.wikimedia.org/glwiki/}, but the differences should have a small effect in practice.} of the Galician version of the Wikipedia. To clean the data, we used \texttt{wikiextractor}\footnote{\url{https://github.com/attardi/wikiextractor}}, which transforms the content of Wikipedia articles into raw text.\footnote{In our work, we kept both the main texts and the headers, too.} We did not apply any further preprocessing steps, in order to do this training phase as self-supervised as possible.
\texttt{Wikiextractor} divides the output into a number of text files of 1MB each. We selected the first 95\% of these Wikipedia articles for the training set (with a total of 42 million tokens), 
and the remainder 5\% for the dev set (2,5 million tokens), 
to keep track of the loss and perplexity at different training points and ensure a successful training. As an enclyclopedic resource widely used by the NLP community, the resulting Wikipedia-based corpus is a well-structured and mostly clean dataset which does not contain as much noise as other crawled corpora (e.g. incomplete sentences, lines with no clear end, etc.).

Contrary to the original BERT release and some other monolingual trainings, we simply pre-train on the masked language objective and ignore the next sentence prediction one, since some recent BERT variants have shown that this second objective adds little or no benefit when it comes to fine-tune BERT-based models for downstream tasks \cite{liu2019roberta}, as we will be doing in this paper (see also Section~\ref{sec:experiments}).

\subsection{Models}
We now describe the procedure that we followed to pre-train our BERT models for language modeling, specifying the differences and similarities with respect to the training of other monolingual BERT models. We also will introduce the framework that we will use to fine-tune our models for downstream tasks.

\subsubsection{BERT tokenizer and sub-word vocabulary} 

\begin{table*}[!ht]
\centering
\begin{tabular}{|l|l|}
  \hline\rule{-2pt}{15pt}
  {\bf Model} & {\bf Tokenization} \\
  \hline\rule{-2pt}{10pt}
  mBERT       & Os nos \#\#os amigos dix \#\#éro \#\#nno \#\#s que o cam \#\#iño era este . \\
  Ours        & Os nosos amigos dix \#\#éron \#\#nos que o camiño era este . \\
  \hline
\end{tabular}
\caption{\label{tab:tokens} Tokenization of the sentence \emph{Os nosos amigos dixéronnos que o camiño era este.} (`Our friends told us that this was the way.') by the original mBERT and our model. Following the same output representation as the BERT tokenizer, we use the symbol \#\# to specify a sub-word that is \emph{not the first} sub-word of a split token.}
\end{table*}

The BERT tokenizer splits the words into the so-called sub-word pieces (essentially n-grams of characters, given a word), where the least common words are simply represented by a generic unknown token, UNK. We follow the same setup as in the original English tokenizer, and define for both models a sub-word cased vocabulary of size 30,000. Such size was also set based both on the vocabularies used for BERT models for related linguistic varieties (e.g., Portuguese \cite{souza2019portuguese} or Spanish \cite{CaneteCFP2020}), and on preliminary tests which suggested that this is a good trade-off between the size and the morphological correspondence of the sub-words.

Even though the tokenizer does not explicitly learn morphological information, better sub-words tend to correspond to morphological affixes, i.e., prefixes, suffixes or stems. In this regard, it is worth noting that other authors have used a larger vocabulary size to train BERT models in agglutinative languages, such as Basque \cite{agerri-etal-2020-give} or Finnish \cite{virtanen2019multilingual}. Exploring an optimal vocabulary size for Galician falls out of the scope of this paper, but it might be an interesting future research line. Nevertheless, we show an example sentence tokenized by mBERT and by our model in Table~\ref{tab:tokens}. As it can be seen, mBERT splits the word \emph{dixéronnos} (`they told us') into several sub-words, including `dix', `éro', `nno', and `s'. Except for the first one (`dix') which corresponds to the frequent irregular root of the verb \emph{dicir} (`tell'), the rest of the original token was split without taking into account morphological boundaries. However, the third sub-word identified by our model (`nos') is a correctly split clitic pronoun (masculine plural dative, `to us'), while the second one (`éron') contains the thematic vowel and person morphemes. In addition, mBERT split the sub-word `iño' from \emph{camiño} (`way' or `path'), which in turn can be confused with the very frequent diminutive suffix `iño' (e.g., the diminutive of \emph{carro} --`car'-- is \emph{carriño}), thus potentially involving inadequate representations of the whole word. Finally, the masculine possessive determiner \emph{nosos} (with both plural possessor and possessed) was also split by mBERT (and not by our model, as it is a frequent word), but in this case it could be argued that the mBERT tokenization might not hurt the model, as the second sub-word `os' corresponds to the masculine plural suffix, while the first one (`nos') could be analyzed as the morphological root (even though it also corresponds to a personal pronoun).

\subsubsection{Pre-training for language modeling}
We have trained two models, varying the number of layers: (i) a BERT with 6 transformer layers (Bertinho\textsubscript{\sc small}), and (ii) a BERT with 12 transformer layers (Bertinho\textsubscript{\sc base}), both trained on the Wikipedia corpus and using the tokenizer presented above. Each layer produces hidden representations of size 768, as in the original BERT paper for monolingual models. We use these models to explore the performance of each architecture in several downstream applications, thus being able to analyze the trade-off between the size and complexity of the models and their quality on extrinsic tasks.

With respect to the hyper-parameter settings, we mostly follow the standard pre-training configuration used in the original BERT paper. We use a learning rate of $1\times10^{-4}$ with a linear weight decay of 0.01. We choose Adam as the network weight optimizer \cite{kingma2014adam} with $\epsilon=1\times10^{-8}$. For the masked language objective and given an input sentence of length $n$, we mask randomly a 15\% of the tokens. From those, 80\% of them are replaced by the wildcard symbol [MASK], 10\% of them are changed to a random word from the input vocabulary, and the remaining 10\% are not modified.

With respect to the training process, in the original BERT model the authors train for 1M steps on sequences of 512 tokens and a batch size of 256. This strategy significantly increases the training time, as the self-attention included at each of the transformer layers in the BERT models runs in $\mathcal{O}(n^2)$, where $n$ is the number of input tokens. Alternatively, authors such as \namecite{agerri-etal-2020-give} use a two-phase procedure using also a training batch size of 256. On the first phase, they train on sequences of length 128 during 900\,000 steps. On the second phase, they continue the training considering sequences of length 512 during 100\,000 additional steps. However, even with this more modest training setup, the authors still had access to a few TPUs.

\begin{figure}
    \centering
    \includegraphics[width=0.95\columnwidth]{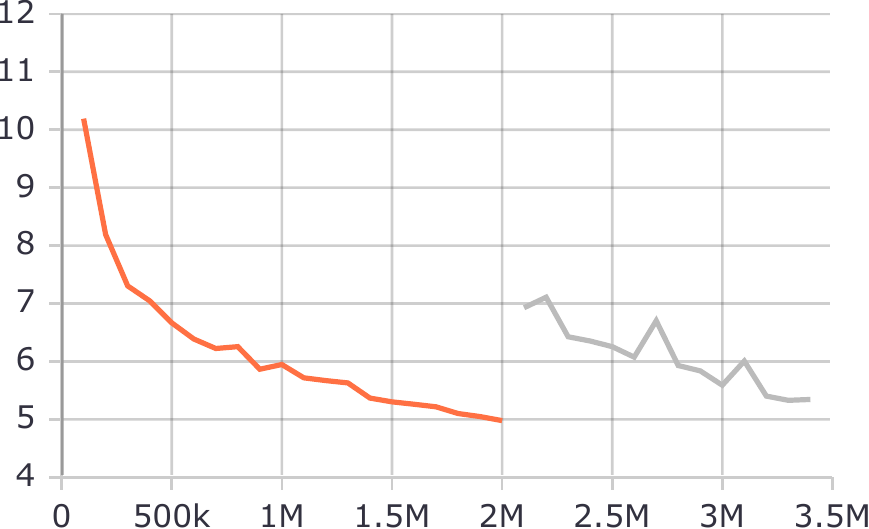}
    \caption{Eval perplexity for Bertinho\textsubscript{\textsc{base}} (12 layers). The eval perplexity over sequences of 128 tokens during the phase 1 of the training is showed in red (left plot). The one over sequences of 512 tokens during the second phase is showed in gray (right plot).}
    \label{fig:eval-perplexity-base}
\end{figure}

\begin{figure}
    \centering
    \includegraphics[width=0.95\columnwidth]{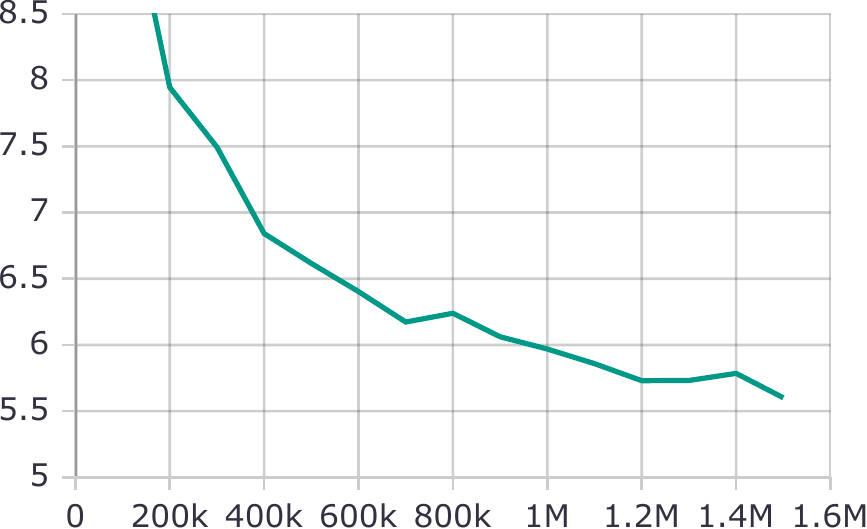}
    \caption{Eval perplexity for Bertinho\textsubscript{\textsc{small}}, trained only over sequence of 128 tokens.}
    \label{fig:eval-perplexity-small}
\end{figure}

In our work, we stuck to a even lower computational resource setup, training the models on a single TESLA P40 GPU of 24GB. Following the standard approach, we first trained the 12-layer model, and consider a two-step training procedure (with the second phase being optional).
For phase 1, we used a smaller training batch imposed by the hardware limitations. More particularly, we used training batches of size 96 considering sequences of 128 tokens. To counteract such smaller training batch, we trained instead the model during more steps, up to 2M. This phase 1 took 30 days to complete the training. Optionally, if the phase 2 was applied, we kept training the model from phase 1 using sequences of length 512. However, this required to limit our training batch size by a significant amount (more particularly, we could only fit 12 sequences in memory). We trained this second sequence for 1.4M additional steps (which took 4 extra days). Thus, for Bertinho\textsubscript{\sc base} we have an additional model trained with the two-phase strategy, that we will be referring as Bertinho\textsubscript{\sc base-2ph}. Figure \ref{fig:eval-perplexity-base} shows the evaluation perplexity obtained at different steps during phase 1 (shorter sequences) and phase 2 (longer sentences).
However, we observed that this second phase was not useful to improve the results over the Bertinho\textsubscript{\sc base} model obtained after finishing phase 1 when it came to fine-tune models for downstream tasks (see also Section \ref{sec:experiments}). 
For the 6-layers model, Bertinho\textsubscript{\sc small}, we decided to apply only the phase 1 of the training procedure, but with batches of size 128 instead and training for 1.5M steps (taking about 22 days to be completed). Figure \ref{fig:eval-perplexity-small} shows the perplexity of the \textsc{small} model during evaluation.

All the models have been trained using the \emph{transformers}\footnote{\url{https://github.com/huggingface/transformers}} library provided by the Hugging Face team \cite{Wolf2019HuggingFacesTS} using, as mentioned, a single 24 GB TESLA P40 GPU.

\subsubsection{Framework}
Given an input sentence $\vec{w}$=$[w_1,w_2,...,w_n]$, we propose to fine-tune the BERT pretrained models for language modeling to a number of tasks involving lexical, syntactic and semantic processing. We will cast these downstream tasks in a sequence labeling setup, i.e, we will learn a mapping $\phi: W^n \rightarrow L^n $ where $L$ will represent the sequence of output labels. The tasks that we will study are POS-tagging, dependency parsing and named  entity recognition, described in detail in Section~\ref{sec:experiments}.

That said, let BERT$(\vec{w})$ be a pre-trained BERT model that maps a one-input vector to a sequence of contextualized representations, $\vec{h}$, we simply map each $h_i$ to an output label $y_i$ through a 1-layer feed-forward network using a softmax ($P(y_i=j|\vec{h}_i)$ = $\mathit{softmax}(\vec{W} \cdot \vec{h}_i + \vec{b})$ = $\frac{e^{\vec{W}_j \cdot \vec{h}_i}}{\sum_{k}^{K}{e^{\vec{W}_k \cdot \vec{h}_i}}}$) to obtain a probability distribution over the set of output labels for each word, $w_i$. In our work, all labels are predicted in an atomic way, that is with single-task learning.\footnote{Especially for more complex tasks, such as dependency parsing as sequence labeling, other authors \cite{strzyz-etal-2019-viable} have decomposed the task into predicting partial labels using a multi-task learning setup.} In all cases, the loss, $\mathcal{L}$, is optimized based on categorical cross-entropy loss ($\mathcal{L} = -\sum{log(P(y_i|\vec{h}_i))}$), and back-propagated through the whole network, i.e. we also fine-tune the BERT weights during the training for the downstream tasks.

\section{Experiments and results\label{sec:experiments}}

To evaluate the usefulness and robustness of the \emph{Bertinho} models when it comes to fine-tuning the model for downstream tasks, we will consider three different problems: (i) part-of-speech tagging, (ii) dependency parsing, and (iii) named entity recognition. We will rely on existing datasets for Galician, which we now briefly review, indicating as well for which tasks we will use them: 

\begin{itemize}

\item \textbf{CTAG corpus:} The \emph{Corpus Técnico do Galego} (Galician Technical Corpus) is a linguistic corpus developed by the Seminario de Lingüística Informática of the University of Vigo, and it is composed of texts from a variety of technical domains (e.g., legal and scientific), totaling almost 18 million words\footnote{\url{http://sli.uvigo.es/CTG/}}. A subset of the CTG corpus, the CTAG (Corpus Técnico Anotado do Galego, `Galician Technical Annotated Corpus')\footnote{\url{http://sli.uvigo.es/CTAG/}}, was manually reviewed \cite{guinovart2009anotacion,agerri-etal-2018-developing}, therefore producing gold-standard resources for POS-tagging and lemmatization (with more than 2 million tokens \cite{ctg}) and NER (with about 200k tokens \cite{sli}). 

\textbf{Tasks and observations:} We will use these resources to evaluate the performance of the BERT-based models on POS-tagging and named entity recognition.

\item \textbf{CTG-UD:} The CTG-UD \cite{gomez2017recursos} is a treebank based on a subpart of the CTG corpus which has been automatically parsed and adapted to Universal Dependencies\footnote{\url{https://universaldependencies.org/}} \cite{nivre-etal-2020-universal}. 

\textbf{Tasks and observations:} CTG-UD will be used to evaluate both POS-tagging and dependency parsing. For POS-tagging, we will consider both universal and language-specific part-of-speech tags (based on the fine-grained tagset of FreeLing \cite{padro11,garcia2010}) as separate tasks. For dependency parsing, and to cast it as a sequence labeling task, we are relying on the encodings proposed by \namecite{strzyz-etal-2019-viable}, which defined different ways to map a sequence of input words of length $n$ into a sequence of syntactic labels, also of length $n$, that can be decoded into full dependency trees. More particularly we will consider their bracketing encoding. Since it is outside of the scope of this work, we leave the details of the encodings for the reader, which can be found in the referrenced work, and we simply use their encoding and decoding functions as a black box.

\item \textbf{Galician-TreeGal:} This treebank is a subset of the XIADA corpus \cite{xiada} that has been annotated following the guidelines of the Universal Dependencies initiative. Galician-TreeGal has around 25k tokens, and its manually revised annotation includes lemmas, POS-tags, morphological features and dependency labels \cite{garcia2016creacion,garcia2018new}. 

\textbf{Tasks and observations:} As in the case of the CTG-UD, we will use this dataset for universal and language-dependent POS-tagging (using the morphologically-rich XIADA tagset)\footnote{\url{http://corpus.cirp.es/xiada/etiquetario/taboa}}, and also for UD dependency parsing. 
It is worth mentioning that due to the small size of the Galician TreeGal treebank, the evaluation can be considered as few-shot learning.

\end{itemize}

All the experiments have been carried out by fine-tuning the BERT-based models (Bertinho\textsubscript{\sc small} and Bertinho\textsubscript{\sc base}) for each specific task using the splits for training and development of each corpus\footnote{Since Galician-TreeGal does not contain a development set, we have splitted the train data into train/dev with a 90\%/10\% ratio. Similarly, we have used 150k tokens from the training set of the CTAG for development of the POS-taggers.} with the referred \textit{transformers} library. As a baseline, we use the official mBERT released by Google (BERT-Base Multilingual Cased, with 12 layers), which includes Galician among its 104 covered languages.

\subsection{Results}
We first show the POS-tagging and NER results on the CTAG corpus, followed by the analyses on the Universal Dependencies treebanks (CTG-UD and Galician-TreeGal).

\subsubsection{CTAG}

\begin{table*}[!ht]
\begin{center}
\begin{tabular} {|l|c|c|c|c|}
 \hline\rule{-2pt}{15pt}
  \multirow{3}{*}{\bf Model}  & \multicolumn{4}{|c|}{\bf CTAG} \\ \cline{2-5}\rule{-4pt}{10pt}
               & \multicolumn{1}{c|}{POS} & \multicolumn{3}{c|}{NER} \\ \cline{2-5}\rule{-4pt}{10pt}
                                            & Acc & P  & R &  F1 \\ 
  \hline\rule{-4pt}{10pt}
  mBERT                                     &93.84 & 83.53  & 85.60  & 84.55 \\
  \hline\rule{-4pt}{10pt}
  Bertinho\textsubscript{\sc small}         &96.16 & 78.40  & 82.68  & 80.48 \\
  Bertinho\textsubscript{\sc base}          &96.40 & 80.49  & 82.74  & 81.60 \\
  Bertinho\textsubscript{\sc base-2ph}      &96.23 & 80.89  & 84.33  & 82.57 \\
  \hline
\end{tabular}
\end{center}
\caption{\label{tab:ctg} POS-tagging and NER (precision, recall, and f-score) results on the CTAG corpus. The POS tagset corresponds to the fine-grained tags of FreeLing (see Section~\ref{sec:data}).}
\end{table*}

Table~\ref{tab:ctg} shows the POS-tagging and NER results on the CTAG corpus. The POS tagset contains 178 tags, and the NER labeling was approached as a BIO classification of four classes (person, location, organization, and miscellaneous) totaling 9 tags (two --B and I-- for each class, and O for the tokens outside the named entities).

On POS-tagging, the best results are obtained by Bertinho\textsubscript{\sc base} (two-phase and single-phase, respectively), followed Bertinho\textsubscript{\sc small}, all of them surpassing mBERT in this task. 
Interestingly, even the 6-layer model obtained better results (2.32\%) than the multilingual one, with 12 transformer layers. However, it is worth noting that mBERT outperformed all the monolingual models on named entity recognition, both in precision and recall, suggesting that for this task multilingual information may improve the performance. These and other results are discussed on Section~\ref{sec:discussion}.

\subsubsection{Universal Dependencies}

\begin{table*}[!ht]
\begin{center}
\begin{tabular} {|l|c|c|c|c|}
  \hline\rule{-2pt}{15pt}
  {\bf Model} & UPOS & FPOS & LAS &UAS  \\
  \hline\rule{-4pt}{10pt}
  mBERT                                     & 95.41 & 91.91 & 76.48 & 80.80\\
  \hline\rule{-4pt}{10pt}
  Bertinho\textsubscript{\sc small}         & 96.42 & 94.56  & 77.59 & 81.55 \\
  Bertinho\textsubscript{\sc base}          & 96.56 & 94.60 & 78.14 & 81.88\\
  Bertinho\textsubscript{\sc base-2ph}      & 96.43 & 94.50 & 77.70 & 81.65\\
  \hline
\end{tabular}
\end{center}
\caption{\label{tab:ctg-ud}POS-tagging accuracies and dependency parsing results on the CTG-UD treebank. POS includes universal and language-dependent tagsets, while parsing is evaluated using LAS and UAS.}
\end{table*}

With respect to the Universal Dependencies treebanks, the results on the CTG-UD are shown on Table~\ref{tab:ctg-ud}, including POS-tagging (on UD POS-tags and using a fine-grained tagset) and dependency parsing (both LAS and UAS values).

In this case, and also in further experiments, Bertinho\textsubscript{\sc base} obtained the best results in both tasks and settings, followed by Bertinho\textsubscript{\sc base-2ph} and by Bertinho\textsubscript{\sc small} (almost tied) and mBERT. As expected, the performance on POS-tagging is higher when using the UPOS tagset (with 16 elements) than with the FPOS one (194 tags). 
On dependency parsing, all the models follow the same mentioned tendency, with Bertinho\textsubscript{\sc base} achieving the best results: 78.14 and 81.88 (LAS and UAS, respectively).

When moving to the Galician-TreeGal treebank (Table~\ref{tab:treegal}), we see again that the best performance is achieved by Bertinho\textsubscript{\sc base} (single and two-phases, respectively), followed by the \textsc{small} variant, and finally by mBERT. It is worth recalling that this treebank is significantly smaller than the other datasets, therefore the model weights have more influence as the fine-tuning process is shorter. In this regard, the results of all models are lower than those obtained on the CTG-UD dataset, achieving POS-tagging accuracies of up to 96.61\% (UPOS, with 16 tags) and 92.70\% (FPOS, containing a tagset of 237 elements). On dependency parsing, the best results were of 75.26\% and 80.27\% on LAS and UAS, respectively.

\begin{table*}[!ht]
\begin{center}
\begin{tabular} {|l|c|c|c|c|}
  \hline\rule{-2pt}{15pt}
  Model & UPOS & FPOS & LAS &UAS  \\
  \hline\rule{-4pt}{10pt}
  mBERT                                     & 94.27 & 87.67 & 71.54 & 77.67 \\
  \hline\rule{-4pt}{10pt}
  Bertinho\textsubscript{\sc small}         & 96.38 & 92.05 & 73.23 & 78.71 \\
  Bertinho\textsubscript{\sc base}         & 96.61 & 92.70 & 75.26 & 80.27  \\
  Bertinho\textsubscript{\sc base-2ph}     & 96.46 & 92.69 & 74.41 & 79.64   \\
  \hline
\end{tabular}
\end{center}
\caption{\label{tab:treegal}POS-tagging accuracies and dependency parsing results on the Galician-TreeGal treebank. POS includes universal and language-dependent tagsets, while parsing is evaluated using LAS and UAS.}
\end{table*}

\paragraph{Significance tests:} We applied significance tests for POS-tagging and parsing, to determine whether the proposed monolingual models are actually different than mBERT. For POS-tagging, we applied a t-test that compares the accuracies per sentence obtained by mBERT and each of the monolingual models. All models are significantly different (with $p<0.01$), except Bertinho\textsubscript{\sc base-2ph} on the Galician-TreeGal (UPOS). For parsing, as in \namecite{vilares-gomez-rodriguez-2018-transition}, we used instead the Bikel’s randomized parsing evaluation comparator, a stratified shuffling significance test. The null hypothesis is that the outputs produced by mBERT and any of the monolingual models are produced by similar models and so the scores are equally likely. To refute it, it first measures the difference obtained for a metric by the two models. Then, it shuffles scores of individual sentences between the two models and recalculates the metrics, checking if the difference is less than the original one, which would be an indicator that the outputs generated by the models are significantly different. All models are significantly different (with $p<0.01$). 

\subsection{Discussion\label{sec:discussion}}
There is a clear tendency with respect to the performance of the different models on the downstream tasks. Except for NER (discussed below), Bertinho\textsubscript{\sc base} (12 layers) consistently obtains the best results, following by Bertinho\textsubscript{\sc small} (6 layers) and the official mBERT release (also with 12 layers).

Apart from the models themselves and from task-specific properties, there are two parameters that seem to play an important role when fine-tuning for a particular downstream task: the size of the tagset and the amount of training data. In this regard, the gain obtained by the best models (when compared to those with lower results) is higher when using large tagsets and smaller datasets, thus suggesting that they encode better information which in turn has a stronger impact on the final performance.

We have assessed this finding by observing the differences between the results of our best model (Bertinho\textsubscript{\sc base}) with those of mBERT:
With respect to the size of the tagset, we compared the results on UPOS and FPOS on both Universal Dependencies corpora. The difference between Bertinho\textsubscript{\sc base} and mBERT on UPOS tagging (with 16 tags) was of 1.15 and 2.34 on CTG-UD and Galician-TreeGal, respectively. However, when using FPOS (with 194 and 237 different tags on the mentioned corpora), the gain achieved by Bertinho\textsubscript{\sc base} increased to 2.69 (CTG-UD) and 5.03 (Galician-TreeGal).

Besides the divergences between the UPOS and FPOS scenarios, these results also indicate that the differences are noticeably larger on the Galician-TreeGal than on the CTG-UD dataset. As the former treebank has less than 14k tokens for training (while CTG-UD has $\approx$ 80k), the differences seem to be mainly caused by the disparity on the amount of training data. This tendency is also displayed on dependency parsing, where the LAS/UAS differences between Bertinho\textsubscript{\sc base} and mBERT are of 1.66/1.08 on CTG-UD and of 3.72/2.60 on Galician-TreeGal. Finally, this tendency does not hold
when comparing the POS-tagging results on both CTAG and CTG datasets (2.56 \emph{versus} 2.69 on the CTAG and CTG-UD, respectively). Additionally, in this case the UD variant is 10 times smaller and has a larger dataset (194 \emph{versus} 178), and consequently the results are lower than in the non-UD corpus (94.60 \emph{versus} 96.40).

The mentioned tendency is not followed in the NER results, where the multilingual model achieved impressive performance (surpassing the best results published by \namecite{agerri-etal-2018-developing}). Bertinho\textsubscript{\sc base} beats the {\sc small} variant by 1.12 points, but mBERT overcomes the best monolingual model (Bertinho\textsubscript{\sc base-2ph}) by 1.98 points. At first glance, we could hypothesize that the multilingual model performs better at NER as \emph{enamex} named entities (locations, people, organizations, and miscellaneous entities) are represented interlinguistically, so that the model takes advantage of information from various languages. Although this may affect the results in some way, a careful analysis of the output of mBERT and Bertinho\textsubscript{\sc base} has shown that most errors of the monolingual model came from variation regarding upper and lowercase. Thus, in expressions such as ``Especies máis afectadas polo \emph{Plano de Selado}'' (`Species most affected by the Sealing Plan'), ``O \emph{Código Civil} actual'' (`The current Civil Code'), or ``As \emph{Illas do Sur}'' (`The Southern Islands'), our model classified the expressions in italic as miscellaneous (the fist two) and location (the last one) entities, but they are not labeled in the gold standard dataset.
About the identification of person entities, Bertinho\textsubscript{\sc base} failed in some complex nouns including prepositions (e.g., ``Bernardo Barreiro de Vázquez Varela'' labeled as two entities --separated by `de'-- instead of one), while some organizations named with common nouns were not identified by the NER system (e.g., ``Cachoeira'', `waterfall').\footnote{Some other errors of Bertinho\textsubscript{\sc base} were due to mislabelings in the gold-standard, as \emph{Kiko} in ``Camiño Neocatecumenal de \emph{Kiko Argüelles}'' (\emph{sic}), annotated as I-PER instead of B-PER.} Therefore, it could be interesting to analyze different techniques to deal with these issues in further work.

Finally, it is worth noting that the Bertinho\textsubscript{\sc base} variant trained with a two-phase strategy (Bertinho\textsubscript{\sc base-2ph}) consistently obtained worse performance than the single-phase one, except in the named entity recognition scenario.
Even though more investigation is needed, these results indicate that in our case, and contrary to previous related work, a two-phase training procedure (training with longer sentences during the second phase) was not beneficial, hurting the performance of the model in the downstream tasks. We hypothesize this might be partially due to our limited hardware resources, that imposed us a very small training batch during the second phase of training.

In sum, we have trained and evaluated two BERT models for Galician that obtain better results than the official multilingual model in different downstream tasks. Interestingly, we provide both a 12-layer model (with greater performance), and a 6-layer one which obtains competitive results with a computationally less expensive architecture.

\section{Conclusion\label{sec:conclusions}}
We have trained two monolingual BERT models for Galician (dubbed \emph{Bertinho}), for which we have followed a low-resource approach with less than 45 million tokens of training data in a single GPU. Both models have been evaluated on several downstream tasks, namely on dependency parsing, NER, and POS-tagging with different tagsets. Moreover, we have shown how a dedicated tokenizer improves the morphological segmentation of Galician words.

Our best model (Bertinho\textsubscript{\sc base}, with 12 layers) outperforms the official BERT multilingual model (mBERT) in most downstream tasks and settings, and even the small one (Bertinho\textsubscript{\sc small} with 6 layers) achieves better results than mBERT on the same tasks. However, it is worth noting that mBERT has worked better than the monolingual models on NER. 
Finally, our experiments have also shown that a two-phase training procedure for language modeling (with more learning steps and training with longer sequence at the end) consistently hurts the performance of the 12-layer model on most scenarios in our setup. We believe this might be due to our limited hardware resources, that forced us to use a small training batch when pre-training with very long sequences.

In further work we plan to carry out a deeper analysis of the NER results, and also to compare the different layers of the BERT models with static word embeddings such as \textit{word2vec} or GloVe. Furthermore, we aim to extend our models using larger corpora. \emph{Bertinho} has been trained on less data than other BERT models for related languages such as BETO \cite{CaneteCFP2020}. We believe collecting more data for other sources such as CommonCrawl could improve the performance.
However it is also fair to state that there are also studies that point in the opposite direction. For instance, the results of \namecite{raffel2020exploring} indicate that smaller clean datasets are better than large noisy corpora, so that it could be interesting to assess to what extent our results (obtained with a small dataset) can be improved with new data crawled from the web \cite{agerri-etal-2018-developing,wenzek-etal-2020-ccnet}.

Finally, it is important to recall that the work performed in this study contributes to the NLP community with the release of two freely available \emph{Bertinho} models for Galician.

\section*{Acknowledgements}

This work has received funding from the European Research Council (ERC), which has funded this research under the European Union's Horizon 2020 research and innovation programme (FASTPARSE, grant agreement No 714150), from MINECO (ANSWER-ASAP, TIN2017-85160-C2-1-R), from Xunta de Galicia (ED431C 2020/11), from Centro de Investigación de Galicia `CITIC', funded by Xunta de Galicia and the European Union (European Regional Development Fund- Galicia 2014-2020 Program), by grant ED431G 2019/01, and by Centro Singular de Investigación en Tecnoloxías Intelixentes (CiTIUS), ERDF 2014-2020: Call ED431G 2019/04. DV is supported by a 2020 Leonardo Grant for Researchers and Cultural Creators from the BBVA Foundation. MG is supported by a Ramón y Cajal grant (RYC2019-028473-I).

\bibliographystyle{fullname}
\bibliography{EjemploARTsepln,anthology,bib.bib}

\end{document}